\documentclass{article}

\usepackage[preprint, nonatbib]{nips_2018}
\usepackage{booktabs} % For formal tables
\usepackage[justification=centering]{caption}
\usepackage{amsmath}
\usepackage{lipsum}
\usepackage{balance}
\usepackage{graphicx}
\usepackage{hyperref}
\usepackage{xcolor}
\usepackage{multirow}

\begin{document}
\title{Neuromodulated Learning in Deep Neural Networks}

\author{Dennis G Wilson\\
University of Toulouse, IRIT - CNRS - UMR5505, Toulouse, France\\
\texttt{dennis.wilson@irit.fr}
\AND Sylvain Cussat-Blanc\\
University of Toulouse, IRIT - CNRS - UMR5505, Toulouse, France\\
\texttt{sylvain.cussat-blanc@irit.fr}
\AND Herv\'e Luga\\
University of Toulouse, IRIT - CNRS - UMR5505, Toulouse, France\\
\texttt{herve.luga@irit.fr}
\AND Kyle Harrington\\
Virtual Technology + Design, University of Idaho, Moscow, ID 83844, USA\\
\texttt{kharrington@uidaho.edu}
}

%% \author{Dennis G Wilson\inst{1} \and Kyle Harrington\inst{2} \and Sylvain
%%   Cussat-Blanc\inst{1} \and Herv\'e Luga\inst{1}}

%% \authorrunning{DG Wilson et al.}
%% \institute{University of Toulouse, IRIT - CNRS - UMR5505, Toulouse, France
%%   \email{\{dennis.wilson, sylvain.cussat-blanc, herve.luga\}@irit.fr} \and
%%   Virtual Technology + Design, University of Idaho, Moscow, ID 83844, USA
%%   \email{kharrington@uidaho.edu}}

\maketitle

\begin{abstract}
  In the brain, learning signals change over time and synaptic location, and are
applied based on the learning history at the synapse, in the complex process of
neuromodulation. Learning in artificial neural networks, on the other hand, is
shaped by hyper-parameters set before learning starts, which remain static
throughout learning, and which are uniform for the entire network. In this work,
we propose a method of deep artificial neuromodulation which applies the
concepts of biological neuromodulation to stochastic gradient descent. Evolved
neuromodulatory dynamics modify learning parameters at each layer in a deep
neural network over the course of the network's training. We show that the same
neuromodulatory dynamics can be applied to different models and can scale to new
problems not encountered during evolution. Finally, we examine the evolved
neuromodulation, showing that evolution found dynamic, location-specific
learning strategies.

%% \keywords{Gene-regulatory networks, Genetic algorithms, Differentiable Learning}
\end{abstract}

\section{Introduction} %SCB ADDED SECTION TITLE
In deep learning and machine learning, the learning process is most often cast
as an optimization problem of the neural network parameters $\theta$, being
synaptic weights and neuron biases, according to some loss function $Q$. In
classification, for example, this loss function can be the mean squared error
between a target classes, $h_i$, and the class given by the deep NN, $X(\theta,
i)$:

\begin{gather}
  Q_i(\theta) = (X(\theta, i) - h_i)^2\\
  Q(\theta) = \frac{1}{n}\sum_{i=1}^n Q_i(\theta)
\end{gather}

The standard approach to optimizing the weights $\theta$ of the ANN is to use
gradient descent over batches of the data. Classic stochastic gradient descent
(SGD) uses a learning rate hyper-parameter, $\eta$ to determine the speed at
which weights change based on the loss function. This method can be improved
with the addition of momentum \cite{nesterov_method_1983}, which changes the
weight update based on the previous weight update. An additional
hyper-parameter, $\alpha$, is then used to determine the
impact of momentum on the final update:

\begin{gather}
  \Delta \theta^{(t+1)} \leftarrow \alpha\Delta \theta^{(t)}-\eta\nabla Q_i (\theta^{(t)})\\
  \theta^{(t+1)} \leftarrow \theta^{(t)} + \Delta \theta^{(t+1)}
  \label{eqn:learning_sgd}
\end{gather}

In contemporary deep learning, there is a variety of gradient descent approaches
to choose from. Adagrad \cite{duchi_adaptive_2011} implements an adaptive
learning rate and is often used for sparse datasets.
Adadelta \cite{zeiler_adadelta:_2012} and RMSprop \cite{tieleman_divide_2018}
were both suggested to solve a problem of quickly diminishing learning rates in
Adagrad and are now popular choices for timeseries tasks.
Adam \cite{kingma_adam:_2014} is one of the most widely used optimizers for
classification tasks, where past gradients are stored in a variable $m$, and
past squared gradients are stored in a variable $v$. Two hyper-parameters,
$\beta_1$ and $\beta_2$, control the update rate of $m$ and $v$, respectively.
$m$ and $v$ are then used to update the weights, instead of using the gradient
directly. This update has a learning rate hyper-parameter, $\eta$, as well as a
``fuzzing factor'' hyper-parameter $\epsilon$ which controls the ratio between
$m$ and $v$ in the final update.

%% \begin{gather}
%%   m_\theta^{(t+1)} \leftarrow \beta_1 m_\theta^{(t)} + (1-\beta_1) \nabla Q_i(\theta^{(t)})\\
%%   v_\theta^{(t+1)} \leftarrow \beta_2 v_\theta^{(t)} + (1-\beta_2)(\nabla Q_i(\theta^{(t)}))^2\\
%%   \hat{m}_\theta = \frac{m_\theta^{(t+1)}}{1-\beta_1^t}\\
%%   \hat{v}_\theta = \frac{v_\theta^{(t+1)}}{1-\beta_2^t}\\
%%   \theta^{(t+1)} \leftarrow \theta^{(t)} - \eta\frac{\hat{m}_\theta}{\sqrt{\hat{v}_\theta}+\epsilon}
%% \end{gather}

These methods, as well as others, are all the result of empirical study on
specific problems. An overview of their different benefits and weaknesses is
presented in \cite{ruder_overview_2016}, and the choice of optimizer represents
an important but difficult decision on the part of the human expert, followed by
the equally difficult choice of hyper-parameters for the chosen method. These
choices depend on domain, on the data available, on the architecture of the
network, and on the training resources available. Furthermore, the choice is
restrained to these existing methods, or to the rigorous development of a new
optimization method.

In this work, we propose a method to automatically develop an optimizer. Using
evolution, a neuromodulatory agent is generated for a training task. This agent,
based on artificial gene regulatory networks (AGRN), uses an existing optimizer
as a base and modifies the parameters of learning at each layer and at each
update. Two different base optimizers were tested: SGD and Adam. We therefore
denote the neuromodulatory versions of these optimizers Nm-SGD and Nm-Adam.
These optimization bases were chosen based on their popularity for the chosen
task, image classification. We use classification on the CIFAR benchmark to
demonstrate this method, but it can be applied to any domain, as evolution can
create an optimizer specialized for the domain of interest. We show that the
evolved neuromodulation strategy can generalize during evolution to different
deep ANN architectures and after evolution to a longer training time and to new
problems. Furthermore, by analysing the behavior of the evolved AGRN during
training, we demonstrate that the location-specific and time-dependent qualities
of neuromodulation are important for deep learning training, as they are in the
biological brain. This represents a novel foray into location-specific learning
for deep neural networks.

\section{Background}

The field of artificial neuromodulation is in rather new, especially when
applied to ANNs. Much of the existing work in this field has focused on
reward-based modulation for unsupervised Hebbian learning to allow for
supervised learning. In the case of Spike-timing Dependent Plasticity, a type of
Hebbian learning based on biological synaptic change, reward-modulated learning
has been used in a variety of tasks \cite{fremaux_neuromodulated_2016}. In
\cite{farries_reinforcement_2007}, a spiking neural network was trained using
neuromodulated STDP to elicit specific spike train. Similarly, in
\cite{izhikevich_solving_2007}, a population response of neurons firing in a
specific group was demonstrated, using a model of neuromodulation based on a
chemical dopamine signal.

A non-spiking example of neuromodulated Hebbian learning is in in
\cite{velez_diffusion-based_2017}, where diffusion-based neuromodulation is used
to eliminate catastrophic forgetting in ANNs. The ANNs used in this work are
shallow networks trained using Hebbian learning to perform multiple tasks.

Artificial neuromodulation has also been studied in other contexts than ANNs. In
\cite{harrington_robot_2013}, a robot agent learns to cover an area in a
reinforcement learning scheme. The agent is trained by SARSA, which is itself
controlled by an evolved neuromodulator. This work was extended in
\cite{cussat-blanc_genetically-regulated_2015}, where evolved neuromodulation
agents facilitate multi-task learning. This work also uses SARSA as the agent.

The task of artificial neuromodulation can be seen as part of the ``learning to
learn'' or ``meta-learning'' problem of optimizing the learning process. Active
learning optimization of this nature has been the topic of much study, such as
in \cite{andrychowicz_learning_2016}, where a secondary ANN is introduced to
optimize the learning of the primary network. In
\cite{miconi_differentiable_2018}, additional Hebbian learning is shown to
improve the performance of networks trained through gradient descent.

Our focus in this work is to evolve the rules of neuromodulation, using two
existing learning methods, SGD and Adam, as a base and improving upon these
methods with evolution. The physical aspects of the neural network are central
to the design of the neuromodulation controller. SGD is traditionally a global
process, with a static learning rate used for all neurons. While methods like
\cite{kingma_adam:_2014} offer dynamic learning rates, these rates are applied
throughout the network. In the brain, the location of a neuron highly impacts
its learning. Neuromodulation often employs volume transmission as a means of
communicating, using chemicals to distribute information as opposed to synaptic
or wired transmission \cite{agnati_understanding_2010}. Clusters of neurons
physically concentrated together will share chemical signals, even if they have
disparate connections. Some neuromodulatory signals are wide-ranged, effecting
entire sections of neurons together, while others are highly localized,
modulating learning in only a few neurons.

%% The neuromodulatory system in the mammalian brain is organized such that three
%% principles are exhibited: (a) many neurons and networks are multiply modulated,
%% (b) there is extensive convergence and divergence in modulator action, and (c)
%% some modulators may be released extrinsically to the modulated circuit, while
%% others may be released by some of the circuit neurons themselves, and act
%% intrinsically \cite{marder_cellular_2002}. These principles were used in the
%% design of these artificial neuromodulation systems.

The neuromodulation controllers were also designed to be responsive, changing
over time and in response to neural activity. In the brain, this is an important
aspect of learning, where neuromodulation is itself controlled by activity in
the brain. In \cite{pignatelli_role_2015}, the role of synaptic plasticity of
dopaminergic neurons is examined, showing that external pressures such as stress
can change the behavior of these neurons. The response of a synapse to dopamine
changes also over time and depends on the state of the neuron.

The goal of the evolved controllers is therefore to use information about the
neurons, i.e. their placement, activity, and learning history, to modify
learning as it happens and in different ways across the network. For this
purpose, we evolve an artificial controller, an AGRN, to regulate the parameters
of learning at each layer. AGRNs, a model based on biological gene regulatory
networks, have been shown to be capable evolved agents on a number of tasks,
including signal processing, robot navigation, and game playing
\cite{disset_comparison_2017}. In the next section, we present the AGRN model
for neuromodulation.

%% As evolution requires a means of comparison between
%% individuals, the quality of learning must be reduced to a fitness value for
%% selection. In this work, we have focused on the efficiency of learning by
%% allowing all individuals to learn for the same amount of time and using their
%% state at the end of this learning period as the evolutionary fitness. Many other
%% metrics of learning exist, such as the ability to generalize to similar cases
%% based on learning a specific case. However, this work is a novel foray into
%% artificial neuromodulation in deep neural networks, and learning efficiency is
%% the primary objective of much of the research concerning artificial learning.

%% In the following sections, we present these two works on neuromodulation. In the
%% first section, AGRNs are evolved to modify the learning at each layer of a deep
%% neural network, changing the parameters of SGD during learning. This is shown to
%% improve upon standard SGD with optimal parameters and matches the Adam optimizer
%% on CIFAR10, an image recognition dataset. In the second section, based on
%% \cite{wilson_learning_2017}, the parameters of a dopamine-based STDP method are
%% evolved. These parameters control the propagation of the dopaminergic signal and
%% its use in STDP, modifying learning in an artificial creature which is learning
%% to swim in an aquatic environment.

\section{AGRN neuromodulation model}

\begin{figure}[h]
  \centering
  \includegraphics[width=0.8\textwidth]{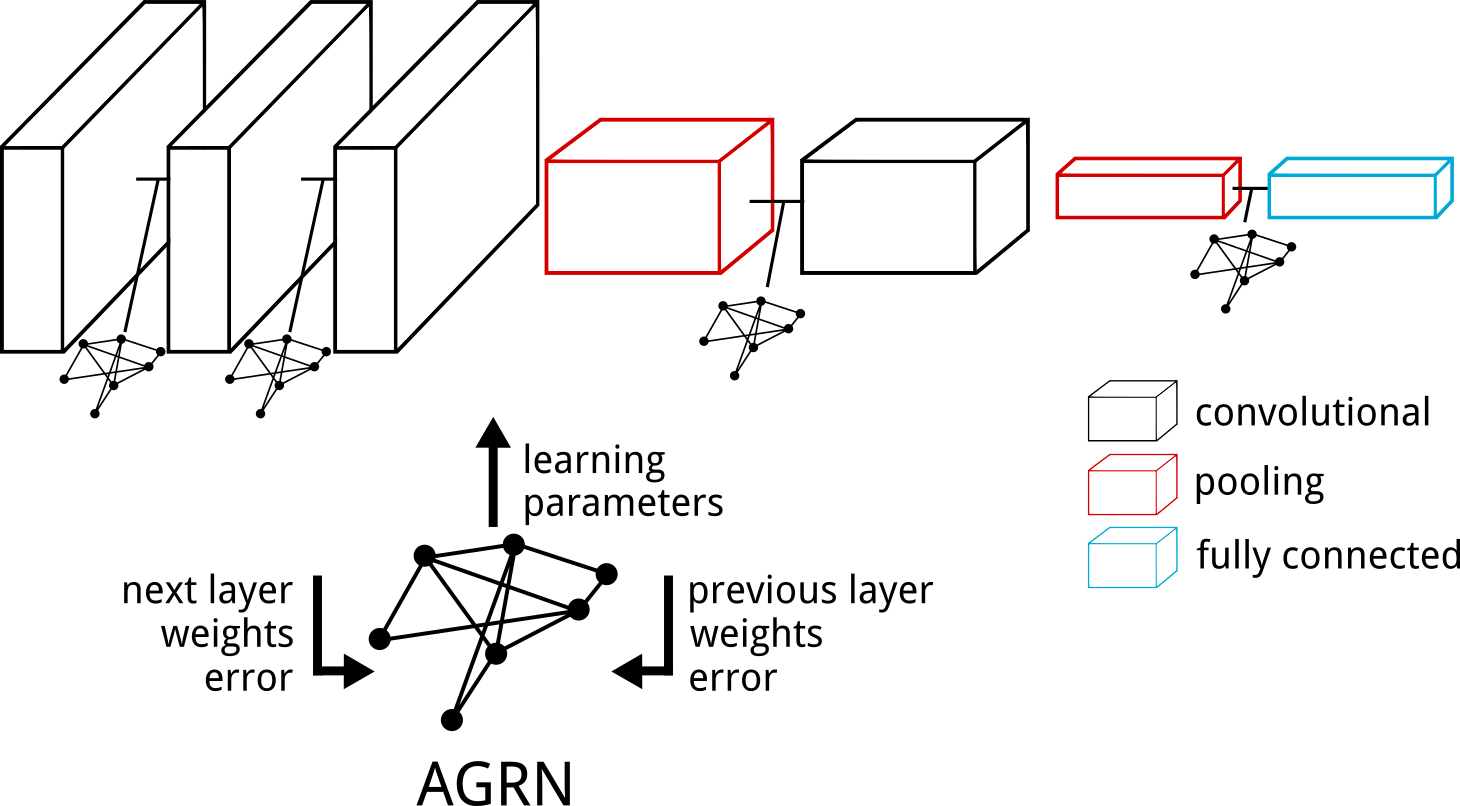}
  \caption{The neuromodulation architecture. A copy of the evolved AGRN is
    placed between all weighted layers of the network. Inputs are given to the
    AGRN at each batch update with information about the two layers between
    which the AGRN is placed. The AGRN then outputs the learning parameters to
    be used in that batch update for the first of the two layers.}
  \label{fig:neuromod_arch}
\end{figure}

The neuromodulation architecture consists of AGRNs placed between all layers of
a deep neural network where weights and gradients are defined (pooling layers,
for example, do not have a corresponding AGRN). The parameters of learning in
the first of the two layers surrounding each AGRN are decided by the AGRN. The
AGRN receives information about the two layers surrounding it, and about the
weights and gradients in both layers. The neuromodulation computation happens in
three steps: 1) collecting the appropriate inputs, 2) processing these through
the AGRN, and 3) using the outputs as learning parameters. This computation
takes place at each update step, i.e. at the end of each batch, which we refer
to as one iteration. In the neuromodulation architecture, each AGRN has the same
genetic code, which is found by evolution; different behavior from the different
AGRN copies arises due to the different inputs given at each layer. A separate
AGRN is used for the synaptic weights and neuron biases of each layer, so there
are two AGRN copies for each layer. All layers except pooling layers use biases.
%% A diagram of the neuromodulation architecture is given in
%% \autoref{fig:neuromod_arch}.

The inputs of each AGRN consist of static information about the layer, i.e. its
location and size, and statistical information about the weights and gradients,
i.e. the mean and standard deviation of both. To be processed by the AGRN, each
input must be between 0.0 and 1.0, so different normalization methods or
constraints are used. For the layer location input, each deep ANN layer is given
a location index $l$ from 0, the first layer, to $L$, the last layer, and the
input is input is $l/L$. The layer size input is similarly normalized over the
entire ANN; the size of each layer, being the number of parameters in the layer,
is divided by the number of parameters of the largest layer in the network. The
statistical information is not normalized but instead constrained in $[0.0,
  1.0]$. The absolute value of the weights are used for the mean and standard
deviation inputs, $\mu_{\theta}$ and $\sigma_{\theta}$, based on the observation
that the magnitude of the weights rarely exceed 1.0. Similarly, the absolute
value of the gradients was used to calculate $\mu_{\nabla Q}$ and
$\sigma_{\nabla Q}$.

These six inputs, layer location, $\mu_{\theta}$, $\sigma_{\theta}$,
$\mu_{\nabla Q}$, $\sigma_{\nabla Q}$, and layer size, are found for each layer.
The AGRN receives the inputs of the layer before and after it, making 12 inputs.
An additional 13th input is also included; this input provides constant
activation of 1.0. This was included to mitigate the possible case that, for
certain layers, none of the inputs would have a high enough magnitude to provide
sufficient activation of the regulatory proteins of the AGRN.

The outputs of the AGRN are the hyper-parameters of the relevant optimizer. For
Nm-SGD, the outputs are the learning rate, $\eta$, and the momentum parameter
$\alpha$. For Nm-Adam, the outputs are the two $\beta$
parameters, $\beta_1$ and $\beta_2$, $\epsilon$, and the learning rate $\eta$. The
full list of inputs and outputs are given in \autoref{tab:neuromod_ins_outs}.

\begin{table}[h]
  \centering
  \begin{tabular}{l|l|l}
    Inputs & SGD output & Adam output\\
    \hline
    layer location & $\eta$ & $\eta$\\
    $\mu_{\theta}$ & $\alpha$ & $\beta_1$\\
    $\sigma_{\theta}$ & & $\beta_2$\\
    $\mu_{\nabla Q}$ & & $\epsilon$\\
    $\sigma_{\nabla Q}$ & &\\
    layer size & &
  \end{tabular}
  \caption{The layer inputs and hyper-parameter outputs of the AGRN. All inputs
    are specific to a layer, and the AGRN receives two copies of these inputs,
    one for each layer surrounding the AGRN. With the constant activation input,
    there are 13 total input proteins. For each output parameter, two AGRN
    output proteins are used. There are therefore 4 output proteins for SGD and
    8 output proteins for Adam.}
  \label{tab:neuromod_ins_outs}
\end{table}

In the standard AGRN update step, protein concentrations are normalized to sum
to 1. This is a part of AGRN computation that has been shown to be necessary
\cite{disset_comparison_2017}, but it can have the undesirable consequence of
restraining the protein concentration levels. In order to allow the AGRNs to
control the magnitude of its outputs, irrespective of normalization, two output
proteins are assigned to each learning hyper-parameter. The normalized
difference between the two output protein concentrations, $o$, is then used to
compute the hyper-parameter output, $O$:

\begin{equation}
  O_i = \frac{|o_{2i}-o_{2i+1}|}{o_{2i}+o_{2i+1}}
\end{equation}

The AGRN used in Nm-SGD therefore has 13 input proteins and 4 output proteins,
and for Nm-Adam it has 13 input proteins and 8 output proteins. These two
different neuromodulation schemes present two different optimization problems;
to find an AGRN, with the respective number of inputs and outputs, capable of
improving overall learning of a deep ANN by making local changes to the learning
parameters at each layer. In the next section, we describe the use of artificial
evolution to find the neuromodulatory AGRNs.

\section{Evolution of the neuromodulatory agent}
\label{sec:neuromod_evolution}

The evolutionary method used in this work is GRNEAT, a genetic algorithm
specialized in AGRN evolution \cite{cussat-blanc_gene_2015}. A key component of
using GRNEAT, or any genetic algorithm, is the design of the evolutionary
fitness function used for selection. In this work, we aim to find an AGRN which
improves learning. Specifically, we use each AGRN individual during training for
the same number of epochs, $E=20$ and then compare these individuals based on
their accuracy on the trained task. In order to ensure generalization, we modify
the deep ANN model and random initialization seed at each generation.

The task used for training during evolution is CIFAR-10
\cite{krizhevsky_learning_2009}. This is a standard image classification task
where 60000 32x32 color images are presented from 10 classes, with 6000 images
per class. We chose this benchmark due to its prevalence in the literature and
for the ease of testing a more difficult problem, CIFAR-100, without needing to
change deep ANN models or the data infrastructure. CIFAR-100 is the same size as
CIFAR-10, but has 100 classes containing 600 images each. Results on the
CIFAR-100 benchmark are presented in \autoref{sec:neuromod_generalization}.

At each generation during evolution, a deep ANN model is chosen. This model is
used to evaluate all individuals in the generation, providing a standard
platform for comparison. We use three different models, $m0$, $m1$, and $m2$,
which are presented in \autoref{tab:neuromod_models}. These models were based on
popular image classification architectures (LeNet and VGG16) and were chosen to
evaluate neuromodulation on a variety of model sizes, from the small $m0$, which
is unable to solve CIFAR-10, to the complex $m2$. The choice of model during
evolution is random; one of the three models is chosen per generation according
to a uniform distribution.

\begin{table}[h]
  \centering
  \begin{tabular}{l|l|l}
    $m0$ & $m1$ & $m2$\\
    \hline
    conv 32 & conv 64 & conv 64\\
    conv 32 & maxpool & maxpool\\
    maxpool & conv 128 & conv 128\\
    conv 64 & maxpool & maxpool\\
    conv 64 & conv 256 & conv 256\\
    maxpool & maxpool & maxpool\\
    fc 512 & conv 512 & conv 512\\
    fc $n_{out}$ & maxpool & maxpool\\
    & fc 4096 & conv 512\\
    & fc 4096 & maxpool\\
    & fc $n_{out}$ & fc 4096\\
    & & fc 4096\\
    & & fc $n_{out}$
  \end{tabular}
  \caption{The three models used during evolution, $m0$, $m1$, and $m2$. Layer
    types are convolutional (conv), maxpool, and fully connected (fc), with the
    size of the layer indicated. $n_{out}$ is the number of outputs and depends
    on problem; for CIFAR-10, used during evolution, $n_{out}=10$.}
  \label{tab:neuromod_models}
\end{table}

Each AGRN individual is therefore used to train a deep ANN, of one of the three
architectures, on the CIFAR-10 dataset, which is split into training and testing
sets of 50000 and 10000 images, respectively. The evolutionary fitness used was
the training accuracy at the end of the final epoch. This metric was chosen to
avoid giving the AGRN an advantage from export to the test data.

Nm-SGD and Nm-Adam agents were evolved using GRNEAT. For the next sections, we
selected the best individuals from 50th generation of each evolution, one for
Nm-SGD and one for Nm-Adam. These two individuals were used to compare
neuromodulation to the base optimization methods in the next section.

\section{Comparison of neuromodulation to standard optimization}

Using the best individual from the last generation of the evolution for Nm-SGD
and Nm-Adam, we compare neuromodulation with standard methods. We first compare
them on the evolutionary task, CIFAR-10. For SGD and Adam, we compare against
the default parameters of the Keras framework\footnote{https://keras.io/}, as
well as parameters found through a grid search. These parameters are presented
in \autoref{tab:neuromod_base_params}, with the ``decay'' parameter being the
learning rate decay over each update.

\begin{table}[h]
  \centering
  \begin{tabular}{*{5}{l}}
    method & model & $\eta$ & $\alpha$ & decay\\
    \hline
    SGD & all & 0.01 & 0.0 & 0.0\\
    \multirow{3}{*}{SGD*} & $m0$ & 0.01 & 0.75 & 0.0\\
    & $m1$ & 0.1 & 0.0 & 0.001\\
    & $m2$ & 0.01 & 0.5 & 0.0\\
  \end{tabular}
  \begin{tabular}{*{7}{l}}
    method & model & $\eta$ & $\beta_1$ & $\beta_2$ & $\epsilon$ & decay\\
    \hline
    Adam & all & 0.001 & 0.9 & 0.999 & $10^-8$ & 0.0\\
    \multirow{3}{*}{Adam*} & $m0$ & 0.001 & 0.9 & 0.999 & 0.001 & 0.0\\
    & $m1$ & 0.1 & 0.99 & 0.9 & 1.0 & 0.001\\
    & $m2$ & 0.1 & 0.99 & 0.999 & 1.0 & 0.001
  \end{tabular}
  \caption{Keras default parameters for the two base methods, and the parameters
    found through search. To find the optimal parameters, all parameter
    combinations over defined ranges were tried, and the best parameter set was
    selected based on final training accuracy on the CIFAR-10 set.}
  \label{tab:neuromod_base_params}
\end{table}

%% These choices are standard for both of
%% these methods, and also standard for image classification tasks. We chose these
%% hyper-parameters due to their prevalence and in order to have a single parameter
%% set across models. Normally, hyper-parameters are optimized for a specific
%% architecture. As we use the same AGRN individual for the three different
%% architectures, we wanted a single hyper-parameter set per optimizer for
%% comparison.

\begin{figure}[h!]
  \centering
  \includegraphics[width=1.0\textwidth]{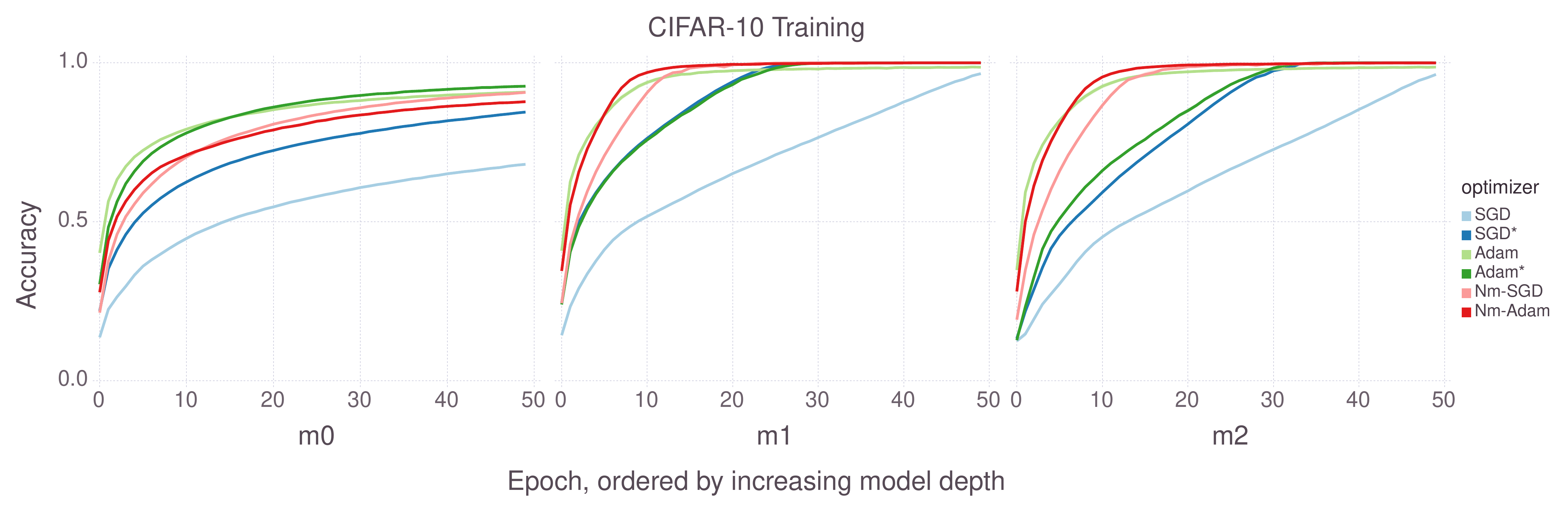}\\
  \includegraphics[width=1.0\textwidth]{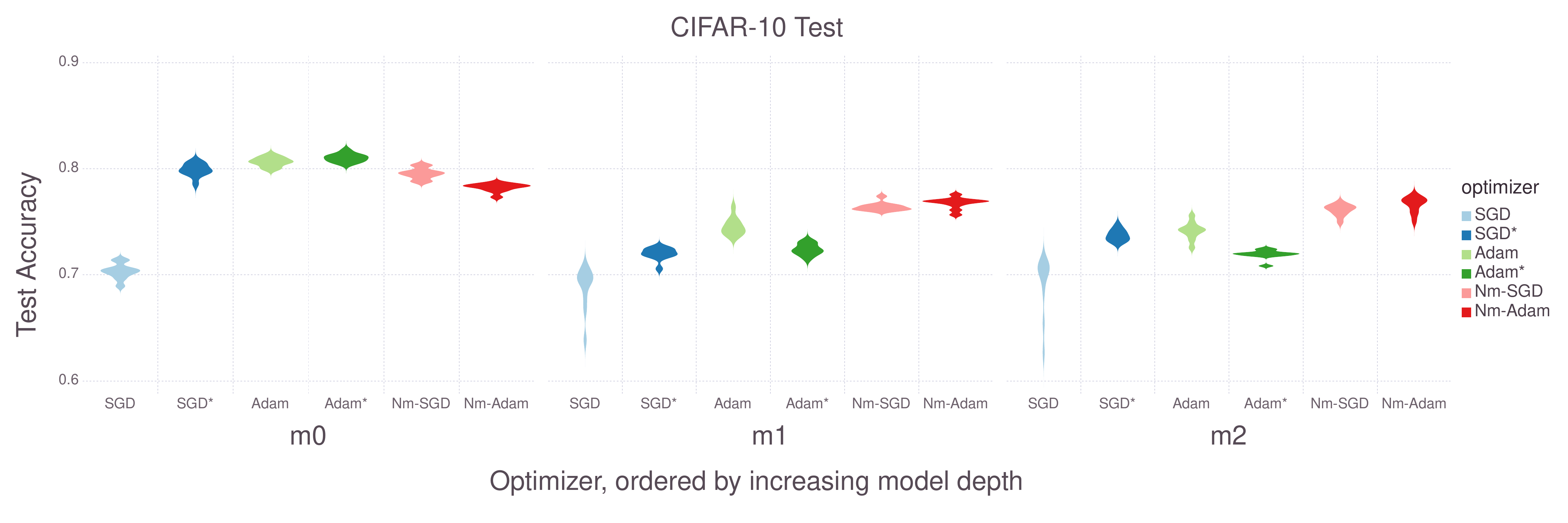}\\
  \caption{Training and test accuracy on the CIFAR-10 benchmark of the
    neuromodulatory methods, Nm-SGD and Nm-Adam, compared to the base methods.}
    %% SGD and Adam, using the default parameters of Keras, and using optimized
  %% parameters (SGD* and Adam*).
  %% Training was performed 10 times on each model
    %% and method. Average training accuracy is shown as a line and one standard deviation
    %% is shown as a ribbon.}
  \label{fig:neuromod_cifar10_fitness}
\end{figure}

We trained the three different models using the compared optimizers 10 times
each. This was done to ensure fair comparison with different random initial
weights. The results of the comparison on training accuracy are presented in
\autoref{fig:neuromod_cifar10_fitness}.
%% , and test accuracy in
%% \autoref{fig:neuromod_cifar10_test_fitness}.

First, we observe that the neuromodulation methods, Nm-SGD and Nm-Adam, are able
to generalize to longer training times. These methods were evolved for 20
epochs, but are here trained for 50 epochs. Especially when training on the
model $m0$, we can see that Nm-SGD and Nm-Adam both continue to aid training
after 20 epochs.

We can also see that neuromodulation can limit overfitting. While the training
accuracy achieved by Nm-SGD and Nm-Adam are both very near 1.0 by the end of 50
epochs on $m1$ and $m2$, the test accuracy for both methods on these models is
superior to that of either standard SGD or Adam.

Finally, we see that the neuromodulatory training can converge faster than
standard training, seen especially for models $m1$ and $m2$. This may be a
result of the use of 20 epochs in the evolutionary fitness, as convergence speed
wasn't explicitly selected for. Instead, by evaluating AGRNs after 20 epochs,
those that converged early may have presented an evolutionary benefit.

%% \begin{figure}[h!]
%%   \centering
%%   \includegraphics[width=1.0\textwidth]{figs/cifar10_sgd_test}\\
%%   \includegraphics[width=1.0\textwidth]{figs/cifar10_adam_test}
%%   \caption{Test accuracy on the CIFAR-10 benchmark of the neuromodulatory
%%     methods, Nm-SGD and Nm-Adam, compared to their base methods, SGD and Adam.}
%%   \label{fig:neuromod_cifar10_test_fitness}
%% \end{figure}

\section{Generalization of the neuromodulatory agent}
\label{sec:neuromod_generalization}

\begin{figure}[h!]
  \centering
  \includegraphics[width=1.0\textwidth]{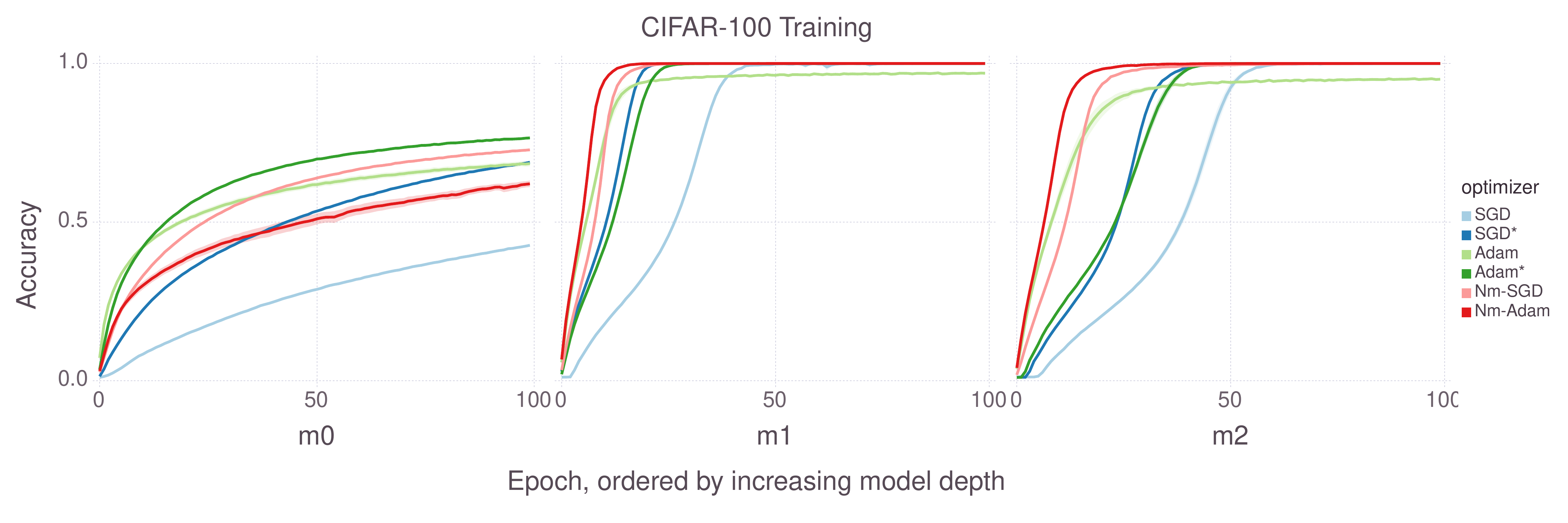}
  \includegraphics[width=1.0\textwidth]{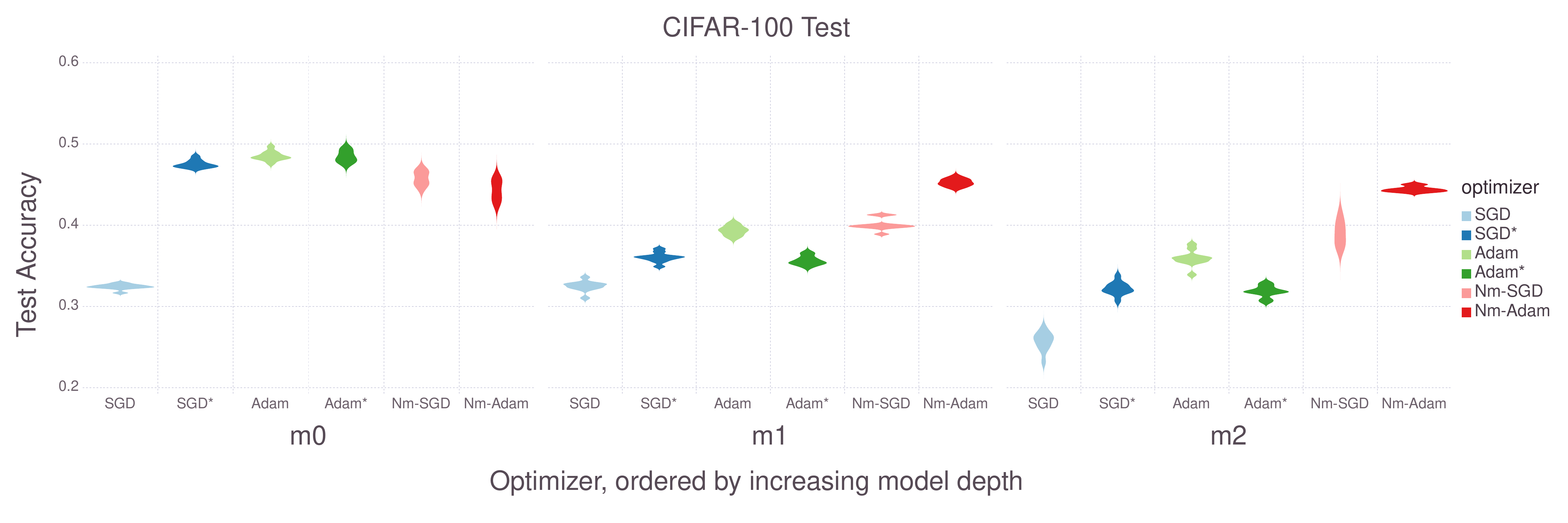}\\
  \caption{Training and test accuracy on the CIFAR-100 benchmark of the
   neuromodulatory methods, Nm-SGD and Nm-Adam, compared to their base methods,
   SGD and Adam.}
  \label{fig:neuromod_cifar100_fitness}
\end{figure}

Next, we compare these four optimization methods on the CIFAR-100 benchmark. The
experiment is the same, i.e. the three different models are trained in 10
separate trials, but the dataset is different and 100 epochs are used to account
for the more difficult dataset. CIFAR-100 is a challenging change from CIFAR-10.
The number of classes increases from 10 to 100, and the number of examples from
each class decreases from 6000 to 600 (5000 to 500 in the training set). The
network must therefore be trained on sparser data for more classes. This new
task would often present the need to find new hyper-parameters, requiring an
expensive tuning search. Here, we use the AGRNs evolved on CIFAR-10 to test
their generalization to CIFAR-100 without change. The results from this
experiment are presented in \autoref{fig:neuromod_cifar100_fitness}.

Nm-SGD and Nm-Adam show clear capabilities to adapt to this new task,
outperforming the base methods on $m1$ and $m2$, while producing early
equivalent test results on $m0$. The training accuracy on $m0$ of Nm-SGD is the
highest of the methods tried, but Nm-Adam is lower than standard Adam. It is
possible that the chosen individual for Nm-Adam was not as suited to the
smallest architecture $m0$ as others from the same evolution, due to the random
selection of model during evolution. Nm-Adam does perform well across the
different models, with better training and test accuracy than the model-specific
Adam* methods.

%% \begin{figure}[h]
%%   \centering
%%   \includegraphics[width=1.0\textwidth]{figs/cifar100_sgd_test}\\
%%   \includegraphics[width=1.0\textwidth]{figs/cifar100_adam_test}
%%   \caption{Test accuracy on the CIFAR-100 benchmark of the neuromodulatory
%%     methods, Nm-SGD and Nm-Adam, compared to their base methods, SGD and Adam.}
%%   \label{fig:neuromod_cifar100_test_fitness}
%% \end{figure}

The conclusions drawn from the CIFAR-10 experiments hold for CIFAR-100. Both
neuromodulation methods exhibit an ability to generalize to longer training,
here extending from 20 epochs during evolution to 100 epochs here. The test
accuracy, especially of Nm-Adam, remains high across different models and is
better than that of SGD and Adam for $m1$ and $m2$. It is worth nothing that the
evolutionary fitness was only related to the training accuracy, and on the
simpler CIFAR-10 set. There was no evolutionary pressure towards generalization
to other problems nor to avoiding overfitting, yet the evolved neuromodulation
dynamics are capable of both.

These experiments demonstrate the viability of this method. Evolved AGRNs can
make effective hyper-parameter choices at each layer, leading to optimized
learning. We now examine the behavior of the AGRNs to understand this function
and the hyper-parameters chosen.

\section{Neuromodulation behavior}

To understand the behavior of the evolved AGRN, we observe the inputs and
outputs of each AGRN copy during training. Specifically, we present the Nm-Adam
CIFAR-10 training of $m1$. This training is shorter than CIFAR-100 and involves
fewer AGRN copies than $m2$. We evaluate a single training, not the training
over 10 different initial weight conditions, as presented in the previous
sections. These choices were made to allow the results to be better understood,
as the inputs and outputs of each AGRN over training represents a large amount
of data, while providing an interesting use-case, the Nm-Adam training on $m1$.
The training is presented in iterations, which are the update steps at each
batch. The batch size used in all experiments was 128, meaning there were 391
iterations per epoch and a total of 19550 iterations over 50 epochs.

\begin{figure}[h!]
  \centering
  \includegraphics[width=0.32\textwidth]{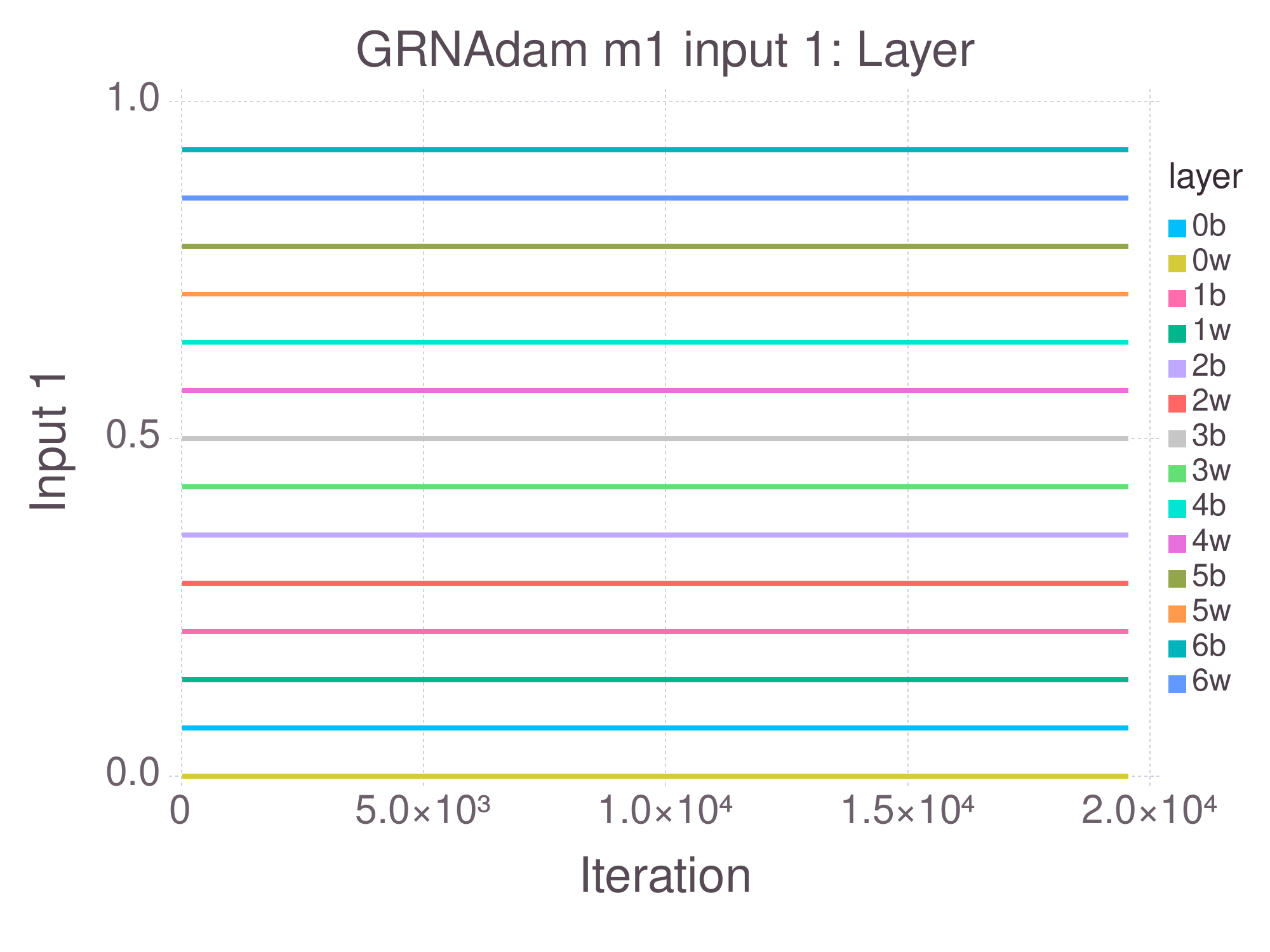}
  \includegraphics[width=0.32\textwidth]{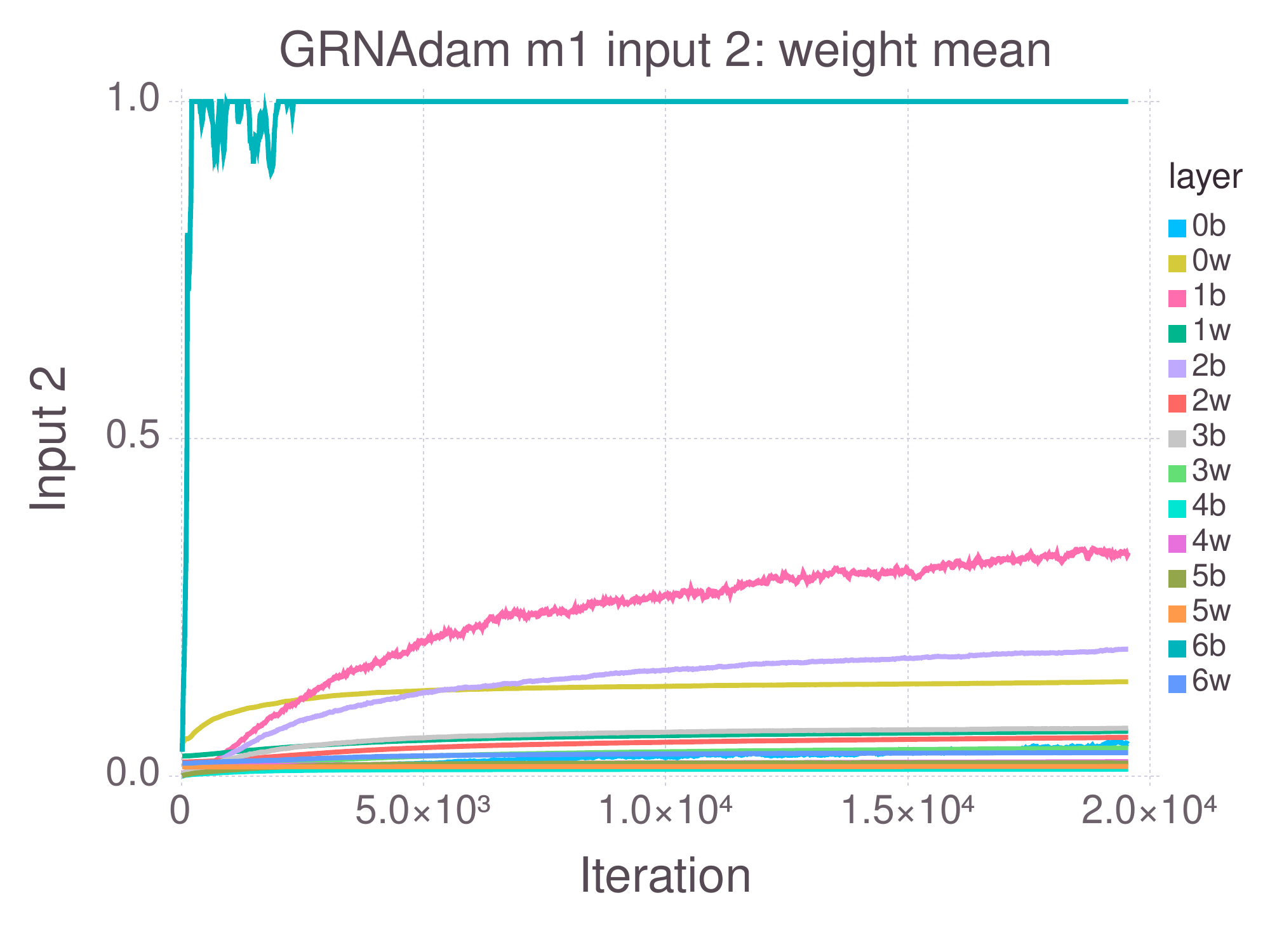}
  \includegraphics[width=0.32\textwidth]{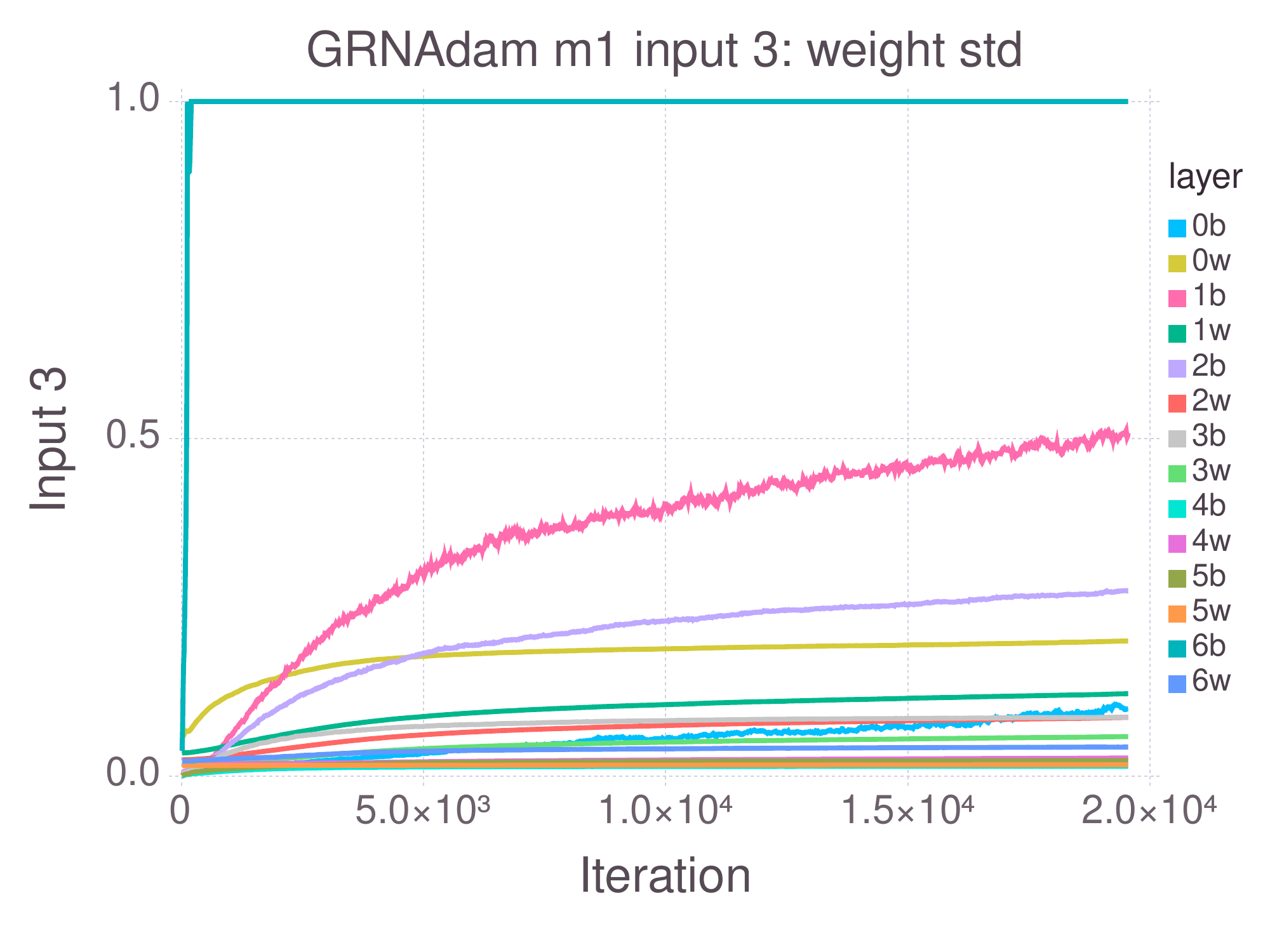}\\
  \includegraphics[width=0.32\textwidth]{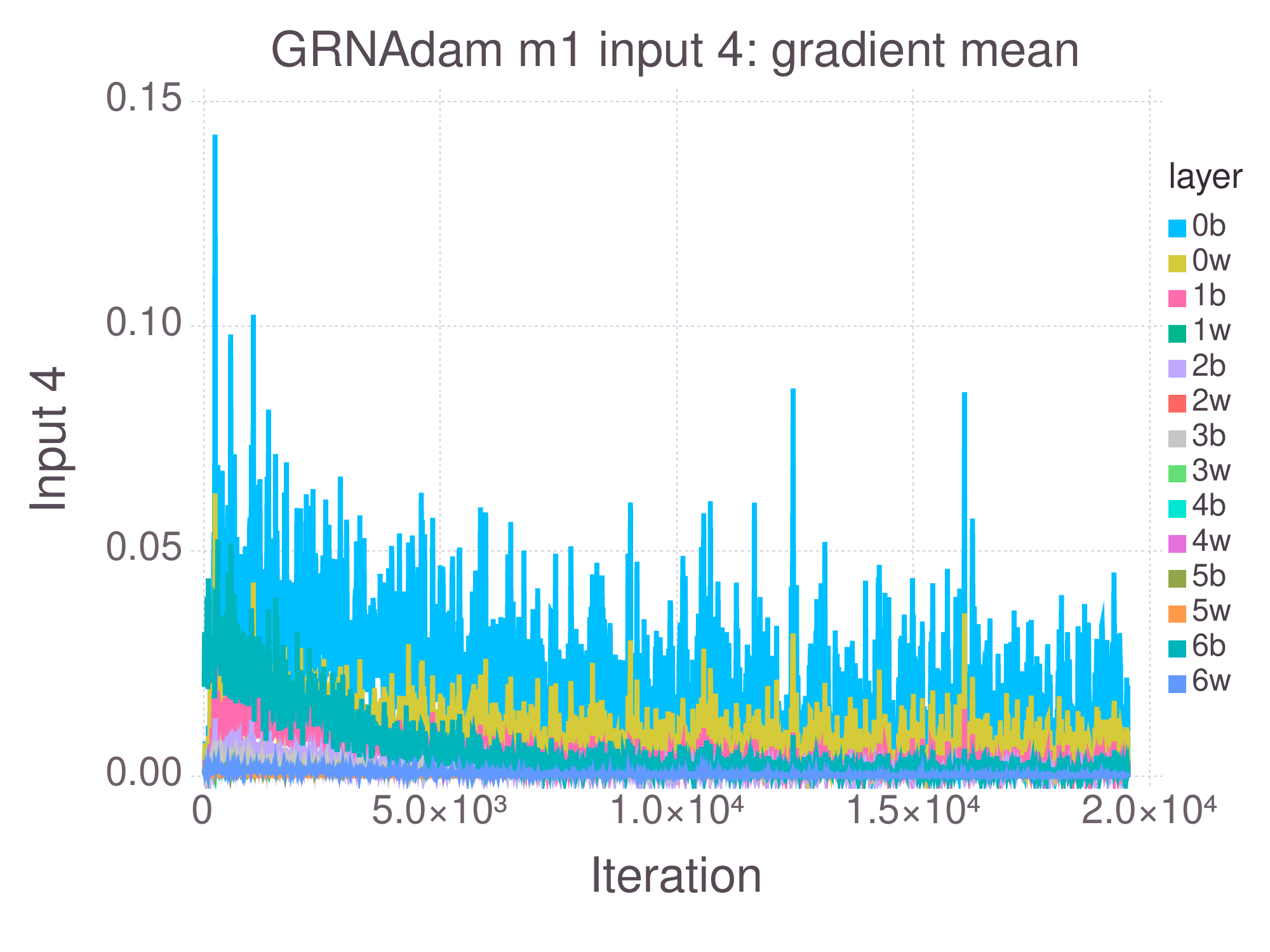}
  \includegraphics[width=0.32\textwidth]{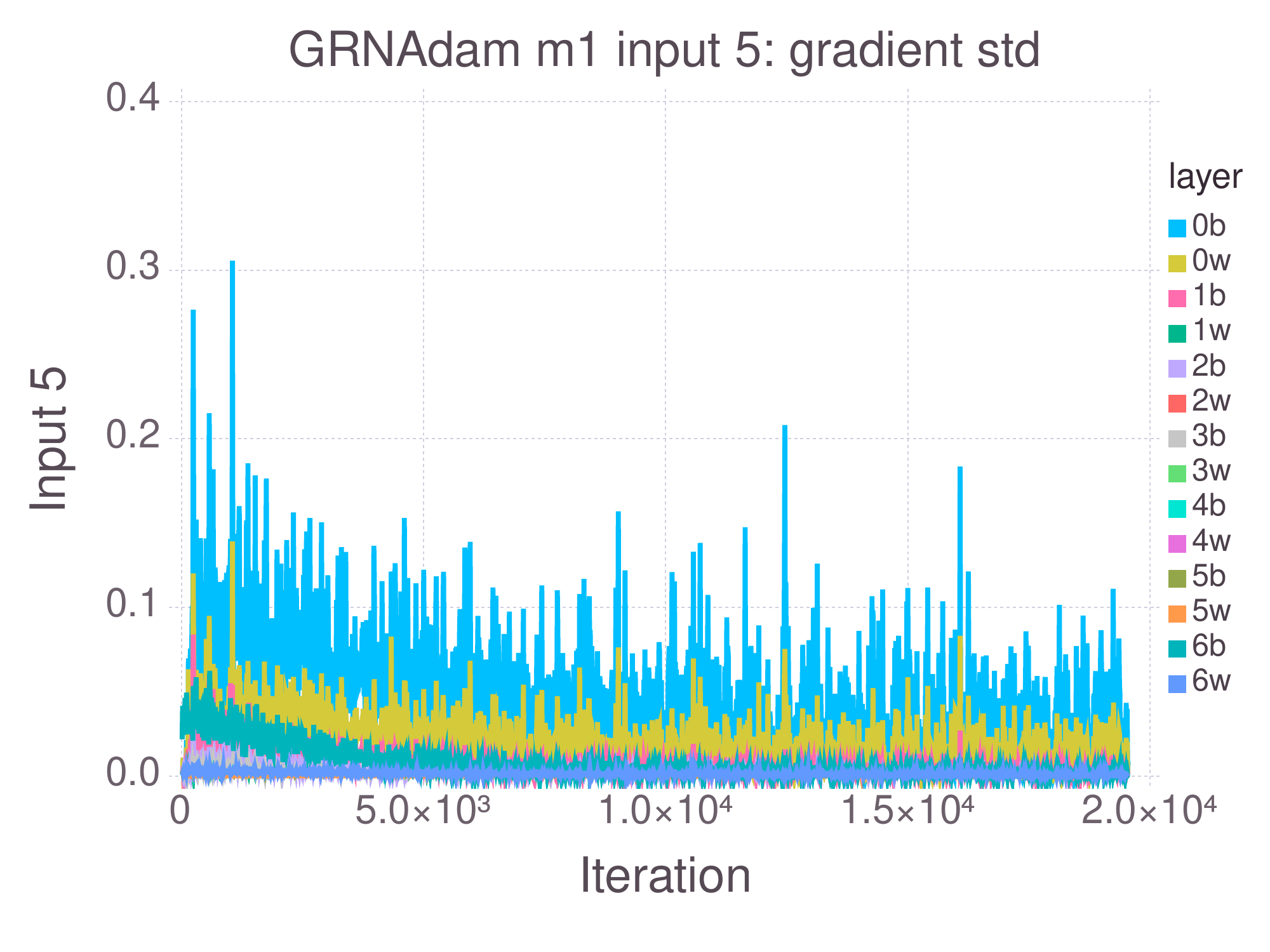}
  \includegraphics[width=0.32\textwidth]{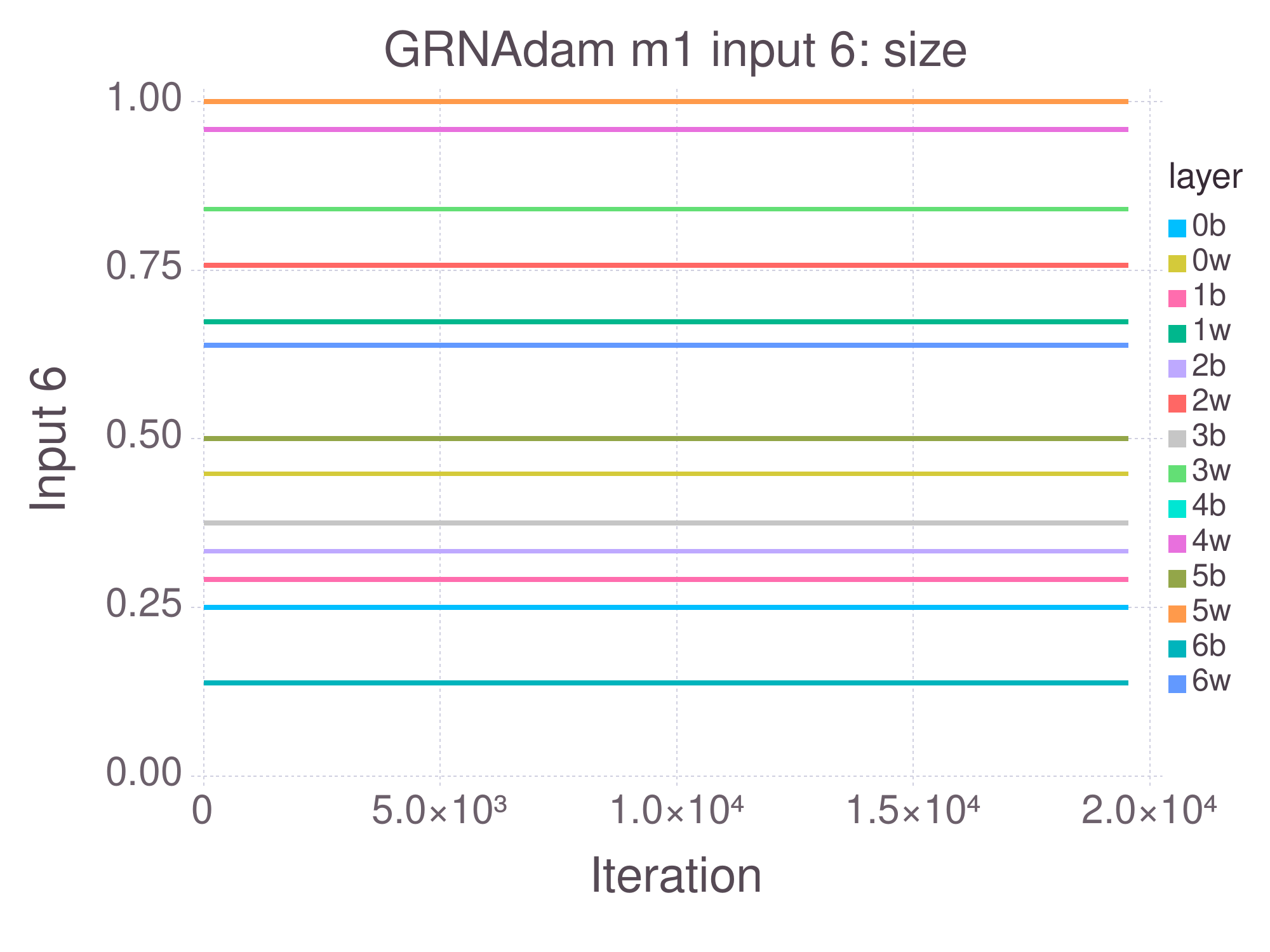}\\
  \caption{Input protein concentrations during Nm-Adam training on $m1$, with
    each AGRN copy represented by a different color. Layer names indicate either
    neuron bias, ``b'', or synaptic weight, ``w''.}
  \label{fig:neuromod_grn_inputs}
\end{figure}

We first present the inputs given to each AGRN in
\autoref{fig:neuromod_grn_inputs}. We show only the first six inputs, as this
contains all of the relevant information. The next six inputs of each AGRN are
the same values for the next layer, already represented in the shown inputs. The
final input, the constant activation input, is always 1.0 for all layers.

The surprising aspect of these inputs is that the biases of the final layer do
exceed 1.0 after a small amount of training. This restricts the information the
AGRN is able to receive about these weights, as $\mu_{\theta}$ is constrained to
1.0. We also see that the gradient mean, $\mu{\nabla Q}$ is generally very small
and could potentially be scaled when provided as an input for easier use by the
AGRN. Finally, while the gradient provides noisy oscillations, most inputs are
static throughout the training, especially after the halfway point of 1e4
iterations.

\begin{figure}[h!]
  \centering
  \includegraphics[width=0.49\textwidth]{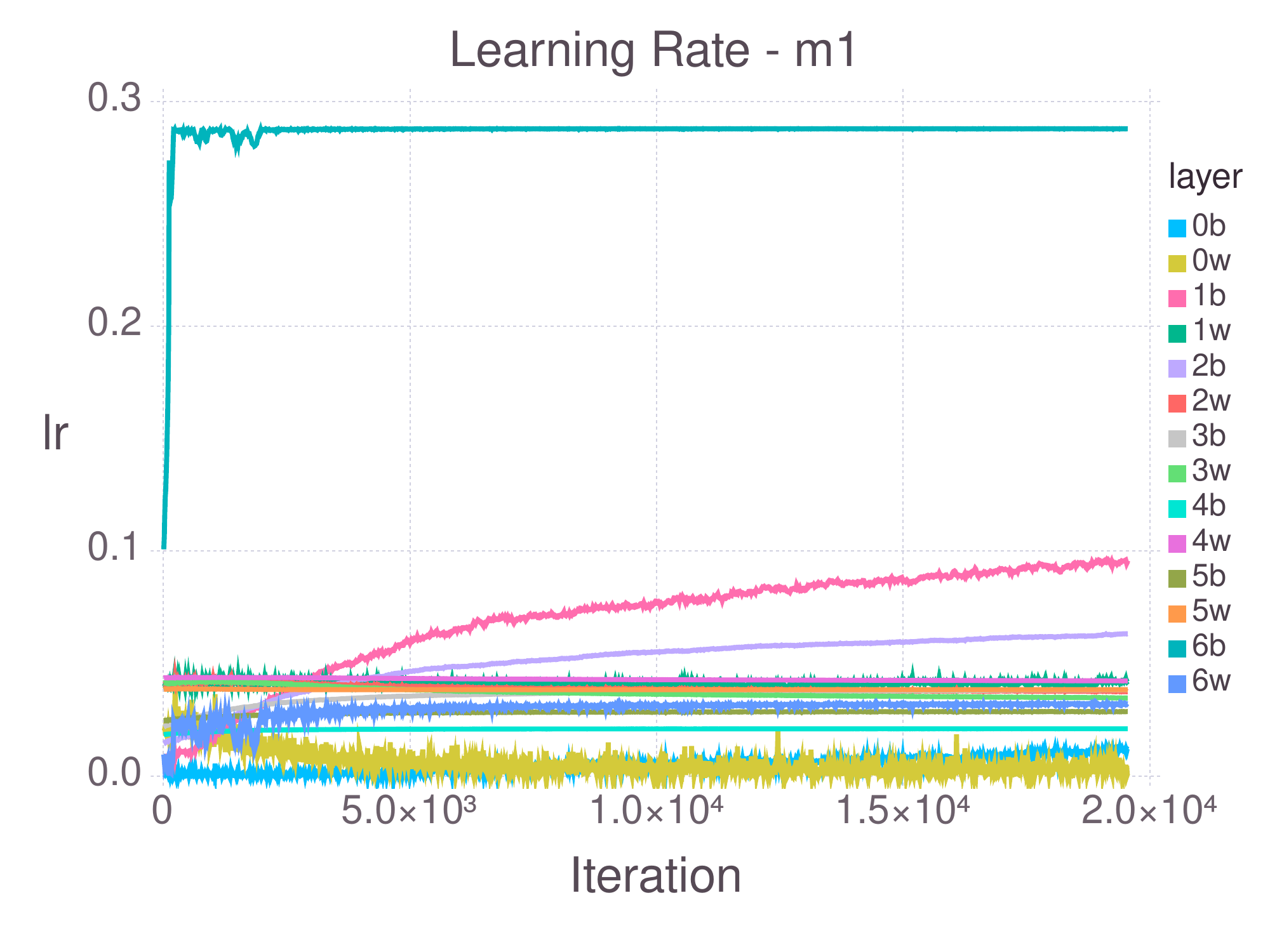}
  \includegraphics[width=0.49\textwidth]{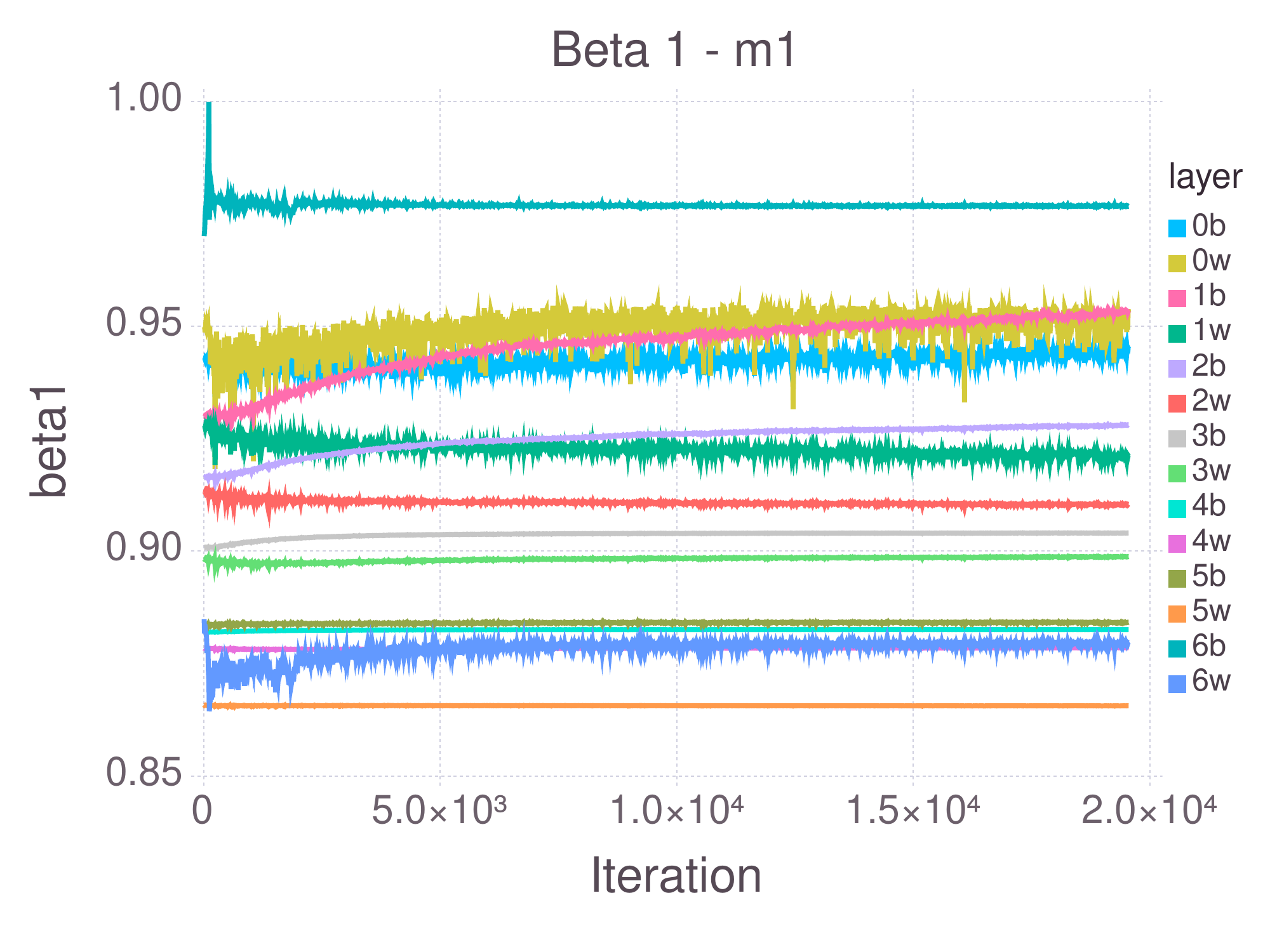}\\
  \includegraphics[width=0.49\textwidth]{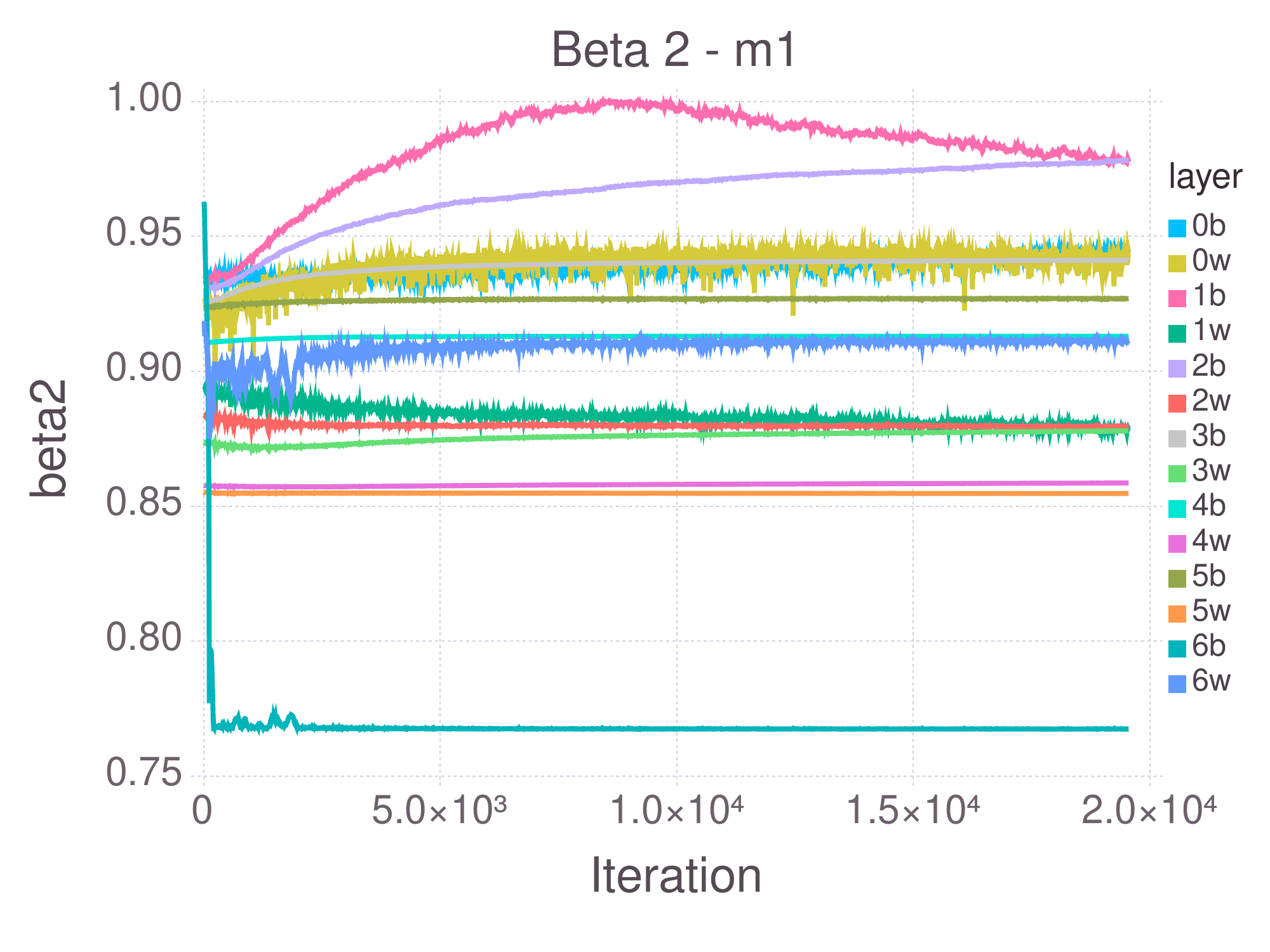}
  \includegraphics[width=0.49\textwidth]{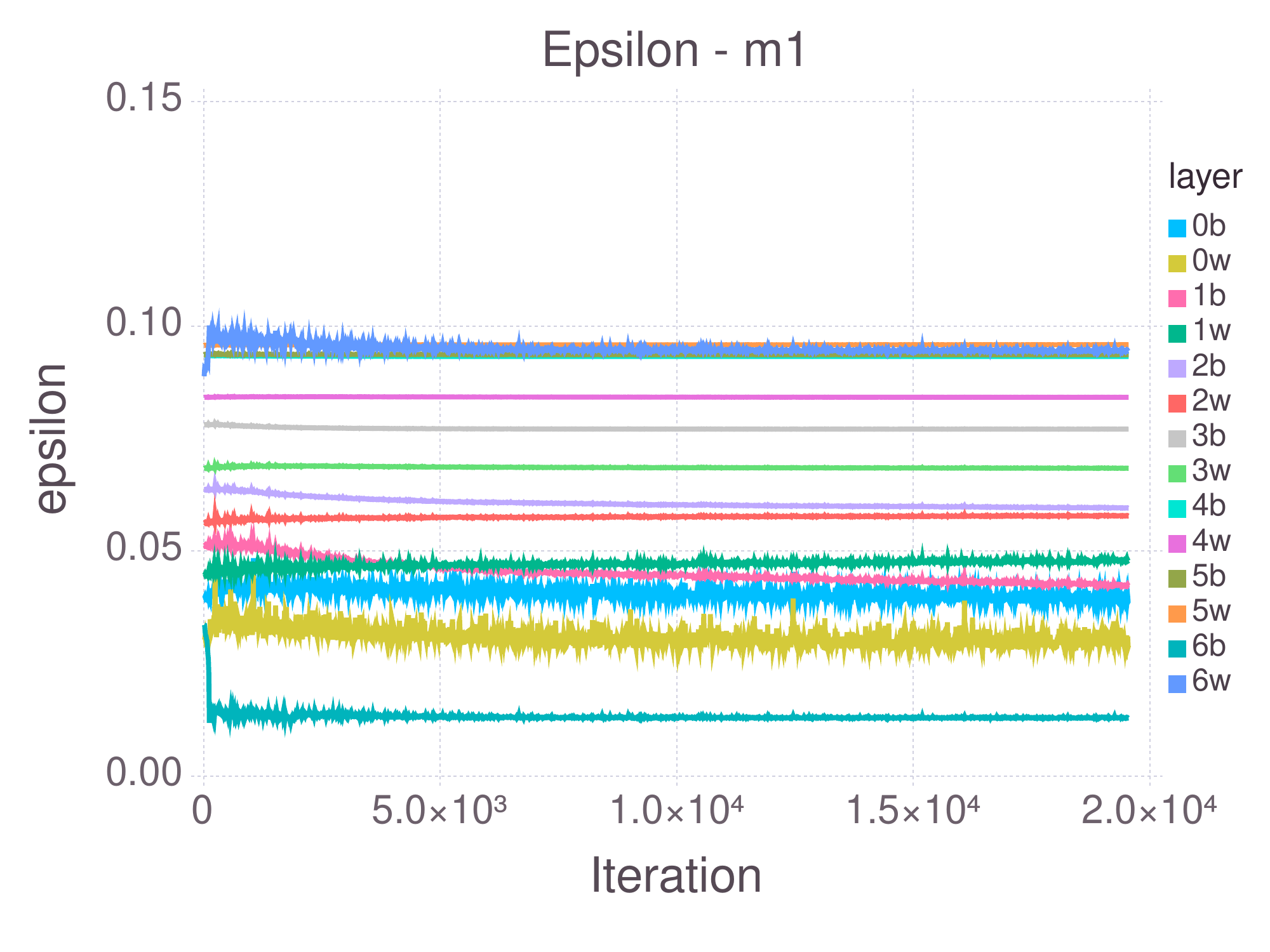}
  \caption{The hyper-parameters chosen during Nm-Adam training on $m1$,
    with each AGRN copy represented by a different color. Layer names indicate
    either neuron bias, ``b'', or synaptic weight, ``w''.}
  \label{fig:neuromod_grn_outputs}
\end{figure}

The hyper-parameters decided by the AGRN are presented in
\autoref{fig:neuromod_grn_outputs}. For this evolved AGRN individual, it is
clear that there is a nearly direct relationship between $\mu_{\theta}$ and
$\eta$ for the biases in the last layer, as the learning rate mirrors the bias
mean almost exactly. There are notable differences in the early layers, however,
with the learning rate of the weights of layer 0 displaying an inverse
relationship with $\mu_{\theta}$. This is an example of a location-specific
neuromodulation strategy, a behavior which is not possible with standard
optimization methods.

Most hyper-parameters are static over time, or like the training, stabilize
after 1e4 iterations. The $\beta_2$ parameter of the biases of layer 1 is an
exception to this, decreasing during the second half of training. This means
that, near the middle of the training, the update of $v_\theta$ depends almost
entirely on its previous state and not on the squared gradient. It only changes
based on the squared gradient at the beginning and ends of training. While this
behavior was not common for the hyper-parameter choices, this sort of
variability over time is a known aspect of neuromodulation and is also not
possible using standard optimization methods.

\section{Conclusion}

In this work, we found that artificially neuromodulated optimizers improved
learning on standard classification benchmarks. Furthermore, neuromodulatory
dynamics were discovered that improve learning on previously unseen problems. As
these dynamics were discovered automatically through evolution, this method of
artificial neuromodulation can be applied to any situation where learning task
performance can be ranked for evolutionary selection.

We believe that this method could lead to shared improvements for deep learning.
At the end of evolution, a single AGRN individual is selected from the entire
population, across all generations. In this work, this was the best individual
from the final population. However, a single evolution creates a wealth of
different individuals, which can be selected based on other characteristics,
like their size. We have shown that a single individual can generalize to longer
training and new problems without needing more training. We can therefore
imagine that neuromodulatory agents could be shared, much as trained neural
networks are shared, with agents evolved for certain task types and
architectures, i.e. image classification. As in the case of ``model zoos'', a
repository of neuromodulatory agents could be shared, eliminating the need to
choose hyper-parameters and improving optimization.

Artificial neuromodulation is a promising direction for deep meta-learning.
Principles of biological neuromodulation were shown in this work to play an
important role in learning. Local signals which change over time were used to
decide the rate of learning and the importance of other factors, such as
momentum, in weight change. These dynamics are novel for stochastic gradient
descent methods, and we believe these characteristics of learning could lead to
the design of new optimization methods.

%% In this work, we have shown a preliminary experiment in the nascent field of
%% artificial neuromodulation. We focused on improving learning and used standard
%% classification benchmarks to evaluate the neuromodulatory agent. However,
%% neuromodulation could assist learning in other ways. For example, catastrophic
%% forgetting is a large problem in deep learning, where ANNs trained on one task
%% completely forget their knowledge when trained on a second task. A
%% neuromodulatory agent could be evolved to mitigate forgetting. In the next
%% section, we focus on reinforcement learning, where neuromodulation acts on a
%% reward signal instead of error gradients. The same could be applied to deep
%% reinforcement learning, where the definition of a gradient is a difficult task.

\section*{Acknowledgments}
This work is supported by ANR-11-LABX-0040-CIMI, within programme ANR-11-IDEX-0002-02.

\bibliographystyle{apalike}
\bibliography{main} 

\end{document}